\documentclass[pdflatex,sn-mathphys-num]{sn-jnl}

\usepackage{graphicx}%
\usepackage{multirow}%
\usepackage{amsmath,amssymb,amsfonts}%
\usepackage{amsthm}%
\usepackage{mathrsfs}%
\usepackage[title]{appendix}%
\usepackage{xcolor}%
\usepackage{textcomp}%
\usepackage{manyfoot}%
\usepackage{booktabs}%
\usepackage{algorithm}%
\usepackage{algorithmicx}%
\usepackage{algpseudocode}%
\usepackage{listings}%
\usepackage{array}%
\usepackage{hyperref}

\begin{document}

\title[Article Title]{Pose-to-Biomechanics: Bridging 3D Human Pose Estimation and Biomechanical Attribute Prediction}


\author{\fnm{Ayda} \sur{Eghbalian}}\email{ayda.eghbalian@utsa.edu}

\author{\fnm{Kevin} \sur{Desai}}\email{kevin.desai@utsa.edu}

\affil{\orgdiv{Department of Computer Science}, \orgname{University of Texas at San Antonio}, \orgaddress{\street{One UTSA Circle}, \city{San Antonio}, \postcode{78249}, \state{Texas}, \country{United States}}}

\abstract{Recent progress in 3D human pose estimation has made markerless recovery of skeletal motion increasingly accurate and scalable. However, most pose estimators remain optimized for geometric keypoint accuracy, while many real-world applications in rehabilitation, sports science, ergonomics, and clinical movement analysis require biomechanical quantities that describe how the body moves, loads, and activates. In this work, we propose \textbf{BioModule}, a lightweight plug-in temporal transformer that attaches downstream of any 3D pose estimator and predicts biomechanical attributes from standard 17-joint 3D skeletons. BioModule is estimator-agnostic and requires no modification of the upstream pose model, enabling existing pose estimators to be extended toward physically interpretable motion analysis.

To train and evaluate BioModule, we construct a large-scale aligned dataset pairing Human3.6M video and 3D keypoints with the biomechanical label space of Human3.6Mplus. We establish and verify anatomical correspondence between coordinate systems of the two datasets, enabling frame-accurate cross-modal supervision. Using this aligned supervision, BioModule predicts biomechanical quantities. We further benchmark BioModule across seven state-of-the-art 3D pose estimators, providing the first systematic analysis of how upstream pose estimation quality propagates to downstream biomechanical prediction fidelity. The results position BioModule as a compact, modular bridge between vision-based pose estimation and biomechanically meaningful human motion analysis.
The complete source code and additional qualitative results are available at:\url{https://utsa-virlab.github.io/BioModule/}}


\maketitle


\section{Introduction}

3D human pose estimation models have become increasingly effective at recovering geometric skeletons from images and videos, yet a gap remains between kinematic pose and the biomechanical quantities required for physically meaningful motion analysis. Existing pose estimation benchmarks commonly optimize for Mean Per Joint Position Error (MPJPE) and related geometric metrics without quantifying physiological correctness, including torques, ground reaction forces, or muscle activation signals, which are essential assets in rehabilitation, and clinical movement assessment.\cite{ionescu2014human36m,martinez2017simple,pavllo20193d,gozlan2025opencapbench,delp2007opensim,seth2018opensim}.

Bridging this gap has traditionally required marker based motion capture laboratories, force plates, and electromyography, instrumentation that is expensive, environment constrained, and difficult to scale to in the wild video. Recent markerless pipelines such as OpenCap~\cite{uhlrich2023opencap} and BioPose~\cite{koleini2025biopose} have demonstrated that video derived kinematics can seed inverse dynamics solvers, but they either require multi view calibrated capture or per subject optimization at inference, limiting their scalability. Meanwhile, body model approaches improve anatomical realism but do not expose the full musculoskeletal output space needed for biomechanical analysis~\cite{keller2023skin,xia2025reconstructing}. The result is that no existing method enables conversion or integration of the 3D skeleton output of standard pose estimations with a comprehensive biomechanical attributes without additional instrumentation.

In this paper, we propose \textbf{BioModule}, a plug-in temporal transformer that attaches downstream of any 3D pose estimator and simultaneously predicts 17~biomechanical criteria across three tiers: kinematic, kinetic, and neuromuscular. These criteria are inferred from a short temporal receptive field of root centered 3D joint positions. The key insight is that a shared temporal encoder, trained on biomechanically labeled skeleton sequences, can learn the implicit mapping from pose dynamics to musculoskeletal state without physics simulation at inference. BioModule introduces no changes to the upstream estimator and requires only a temporally ordered 17-joint 3D skeleton as input, making it compatible with the full spectrum of contemporary lifting models.

To support training and evaluation at scale, we construct Human3.6Mplus, a large-scale dataset that aligns Human3.6M~\cite{ionescu2014human36m} keypoints with musculoskeletal marker-based simulation labels from Human3.6Mplus \cite{nasr2024using}. A core contribution of this dataset is the establishment and geometric verification of anatomical correspondence between the coordinate systems of the two modalities, anchored at the shared pelvis root, enabling frame accurate cross-modal supervision across all frames of the seven subjects and thirty activities included in Human3.6M.

Finally, we conduct a systematic cross-estimator benchmarking of downstream biomechanical prediction quality. Training BioModule once on ground-truth poses and evaluating it, both frozen and after lightweight fine-tuning, across seven state-of-the-art 3D pose estimators, we quantify precisely how upstream pose accuracy propagates to each biomechanical tier. This analysis reveals which architectural families of pose estimator are most biomechanically faithful, providing actionable guidance for practitioners deploying markerless biomechanical pipelines.

\noindent The contributions of this work are:

\begin{itemize}

    \item \textbf{BioModule:} We have trained a plug-in temporal transformer with a tiered multi-head prediction architecture that regresses 17~biomechanical criteria, spanning joint coordinates, torques, ground reaction forces, and muscle activation signals, from any 3D pose estimator's output.

    \item \textbf{Aligned pose and biomechanics dataset:} We have curated a biomechanically annotated dataset pairing Human3.6M video and skeleton sequences with full musculoskeletal simulation labels, including geometric verification of the cross-modal anatomical correspondence required for frame accurate supervision.

    \item \textbf{Cross-estimator biomechanical benchmark:} We have set up an evaluation protocol across seven contemporary 3D pose estimators, providing the first quantitative analysis of how pose estimation architecture and accuracy determine downstream biomechanical prediction fidelity.

\end{itemize}

\section{Related Work}

\subsection{Deep Learning-Based Human Pose Estimation}

Human pose estimation has progressed from direct coordinate regression in single images to increasingly structured models that exploit spatial, temporal, and kinematic priors. Early deep learning approaches such as DeepPose formulated pose estimation as direct regression from image evidence to body-joint locations~\cite{toshev2014deeppose}. Subsequent methods improved localization by using convolutional heatmap representations and multi-stage refinement, including convolutional pose machines~\cite{wei2016cpm}, stacked hourglass networks~\cite{newell2016hourglass}, and high-resolution representations such as HRNet~\cite{sun2019hrnet}. These methods established strong 2D pose detectors that later became the input backbone for many monocular 3D human pose estimation systems.

A major line of 3D pose estimation first estimates 2D joints and then lifts them into 3D space. Martinez et al. showed that a simple fully connected residual network can be highly effective for 2D-to-3D pose lifting when accurate 2D detections are available~\cite{martinez2017simple}. VideoPose3D extended this formulation with temporal convolutional networks, using motion context across frame windows to improve robustness against depth ambiguity and 2D detection noise~\cite{pavllo20193d}. Large-scale motion-capture datasets such as Human3.6M have provided the standard benchmark for evaluating these models in controlled indoor settings~\cite{ionescu2014human36m}.

More recently, transformer-based models have become prominent for video-based 3D pose estimation. PoseFormer introduced a spatial-temporal transformer for modeling joint relations within frames and temporal dependencies across frames~\cite{zheng2021poseformer}, building on the general attention mechanism introduced in Transformer architectures~\cite{vaswani2017attention}. Subsequent methods improved efficiency, ambiguity handling, and motion representation. MHFormer introduced a multi-hypothesis transformer to address monocular depth ambiguity~\cite{li2022mhformer}, PoseFormerV2 used frequency-domain representations for efficient temporal modeling~\cite{zhao2023poseformerv2}, and MotionAGFormer combined graph-based skeletal structure with attention mechanisms~\cite{mehraban2024motionagformer}. More recent models further explore diffusion-based pose aggregation~\cite{shan2023d3dp}, kinematic and trajectory priors~\cite{peng2024ktpformer}, implicit temporal pose proxies~\cite{liu2025tcpformer}, state-space sequence modeling~\cite{huang2025posemamba}, and general human motion representations~\cite{zhu2023motionbert}. Pre-layer normalization has also been shown to improve transformer training stability~\cite{xiong2020layer}.

These methods show that temporal and structural modeling are essential for accurate 3D pose recovery. However, their objective remains primarily geometric: the output is a 3D skeleton or pose representation optimized by joint-position error. Even when such models implicitly encode motion dynamics, they do not directly supervise or evaluate biomechanical quantities such as torques, reaction forces, contact, activation, or excitation. BioModule uses the 3D skeletons produced by such estimators as input, but evaluates them through biomechanical prediction fidelity rather than geometric accuracy alone.

\subsection{Biomechanically Accurate Body Models and Motion Representations}

Beyond sparse keypoints, parametric body models provide richer representations of human shape and pose. SMPL introduced a learned skinned body model that represents human body shape and articulation with a compact parameterization~\cite{loper2015smpl}. Image- and video-based mesh recovery methods such as HMR and VIBE estimate body pose and shape from monocular visual input, enabling temporally coherent reconstruction of human motion in mesh space~\cite{kanazawa2018hmr,kocabas2020vibe}. These methods provide a more complete geometric representation than a sparse 3D skeleton and have become widely used in human motion analysis.

Recent work has further sought to make body models more anatomically and biomechanically meaningful. Keller et al. introduced SKEL and BioAMASS to connect surface body models with a biomechanically grounded skeleton~\cite{keller2023skin}. Xia et al. reconstructed humans with biomechanically accurate skeletons, further emphasizing the importance of anatomical structure in human reconstruction~\cite{xia2025reconstructing}. SKEL-CF extends this direction through coarse-to-fine recovery of biomechanical skeleton and surface mesh representations~\cite{li2025skel}. These works are important because they move beyond visual surface reconstruction toward body representations that are more consistent with human anatomy.

However, body models and mesh recovery methods primarily address how the body is represented or reconstructed. They do not directly provide a general estimator-agnostic mechanism for converting the standard 17-joint outputs of existing 3D pose estimators into a broad biomechanical state space. BioModule is complementary to these approaches: rather than proposing a new body model, it learns to infer kinematic, kinetic, contact, and neuromuscular attributes from the sparse skeleton representation already produced by contemporary 3D pose estimators.

\subsection{Markerless Biomechanics and Video-to-Biomechanics Pipelines}

Biomechanical movement analysis traditionally relies on marker-based motion capture, force plates, and musculoskeletal modeling to estimate kinematics, kinetics, and muscle-related quantities. Recent markerless systems have attempted to reduce this dependence on laboratory instrumentation. OpenCap uses videos from smartphones to estimate human movement kinematics and dynamics through a pipeline that combines pose estimation, musculoskeletal modeling, and simulation~\cite{uhlrich2023opencap}. OpenCap Monocular further extends this direction toward single-video biomechanical analysis~\cite{gilon2026opencap_mono}. OpenCapBench explicitly frames the gap between pose estimation and biomechanics by evaluating whether pose-estimation outputs preserve biomechanically relevant correctness, not only geometric accuracy~\cite{gozlan2025opencapbench}.

Other video-to-biomechanics and markerless motion-capture studies examine different parts of this pipeline. Cotton et al. studied trajectory optimization and inverse kinematics for biomechanical analysis of markerless motion-capture data~\cite{cotton2023markerless}. Ruescas-Nicolau et al. investigated keypoint augmentation for markerless motion capture in biomechanical applications~\cite{ruescasnicolau2024markerless}. Auer et al. evaluated markerless motion capture combined with musculoskeletal models for kinematic analysis~\cite{auer2024markerless}, while Barzyk et al. studied smartphone-based markerless capture of lower-limb joint angles during countermovement jumps~\cite{barzyk2024smartphone}. Rode et al. assessed monocular human pose estimation models for clinical movement analysis~\cite{rode2025clinical}. Together, these studies show that markerless motion capture is increasingly relevant for biomechanics, rehabilitation, sports science, and clinical assessment.

Recent methods also integrate vision models more directly with biomechanical constraints. BioPose estimates biomechanically accurate 3D pose from monocular video by combining mesh recovery with biomechanical constraints~\cite{koleini2025biopose}. Lin et al. used biomechanical models and synthetic training data to estimate 3D kinematics from video~\cite{lin2024kinematics}. Miller et al. showed that integrating machine learning with musculoskeletal simulation can improve OpenCap video-based dynamics estimation~\cite{miller2025integrating}. These approaches demonstrate the value of combining learned visual representations with biomechanical modeling.

Most existing markerless biomechanics systems are designed as complete pipelines involving video processing, pose recovery, trajectory refinement, inverse kinematics, musculoskeletal modeling, or simulation. BioModule addresses a different setting: given the 17-joint 3D skeleton output of an existing pose estimator, it learns a compact temporal mapping to multiple biomechanical attributes. This makes the proposed framework suitable for comparing different upstream pose estimators under a shared biomechanical prediction interface.

\subsection{Musculoskeletal Simulation, Physics-Informed Modeling, and Biomechanical Datasets}

Musculoskeletal simulation provides the physical foundation for estimating biomechanical variables that are not directly visible from video. OpenSim is a widely used framework for creating and analyzing dynamic simulations of movement~\cite{delp2007opensim}, and later extensions support musculoskeletal dynamics and neuromuscular control modeling for human and animal movement~\cite{seth2018opensim}. AddBiomechanics automates model scaling, inverse kinematics, and inverse dynamics from motion-capture data and musculoskeletal models~\cite{werling2023addbiomechanics}. Its associated dataset captures the physics of human motion at scale, providing a broader source of motion and biomechanical supervision~\cite{werling2024addbiomechanics_dataset}.

Human3.6Mplus provides another important form of biomechanical supervision by pairing Human3.6M motion with physically consistent musculoskeletal labels, including kinematic, dynamic, and muscle-related quantities~\cite{nasr2024using}. This type of dataset is critical for learning mappings from pose sequences to biomechanical attributes because quantities such as joint torques, reaction forces, activation, and excitation are not directly annotated in conventional computer-vision pose datasets.

Related work has also explored physics-informed or physics-based learning for pose and dynamics estimation. IPMAN incorporates intuitive physics constraints, including floor contact and center-of-pressure or center-of-mass consistency, into 3D human pose estimation~\cite{tripathi2023ipman}. SSPINNpose uses a self-supervised physics-informed neural network for inertial pose and dynamics estimation~\cite{gambietz2025sspinnpose}, and OrientationNN provides a lightweight physics-informed approach for real-time joint kinematics estimation from IMU data~\cite{bian2025orientationnn}. These methods show that physical constraints can improve motion estimation, although they often target pose, inertial kinematics, or specific dynamics settings rather than a broad simulation-derived biomechanical label space from video-based 3D skeletons.

Overall, musculoskeletal simulation and physics-informed learning provide the label sources and physical constraints needed for biomechanical inference. However, these tools are typically used either as explicit simulation pipelines or as constraints inside task-specific models. BioModule uses simulation-derived supervision differently: it learns a reusable temporal mapping from standard 17-joint 3D pose sequences to multiple biomechanical criteria, enabling biomechanical prediction without running full musculoskeletal simulation at inference time.

\section{Methodology}

\subsection{Overview}
\begin{figure*}[t]
    \centering
    \includegraphics[width=\linewidth]{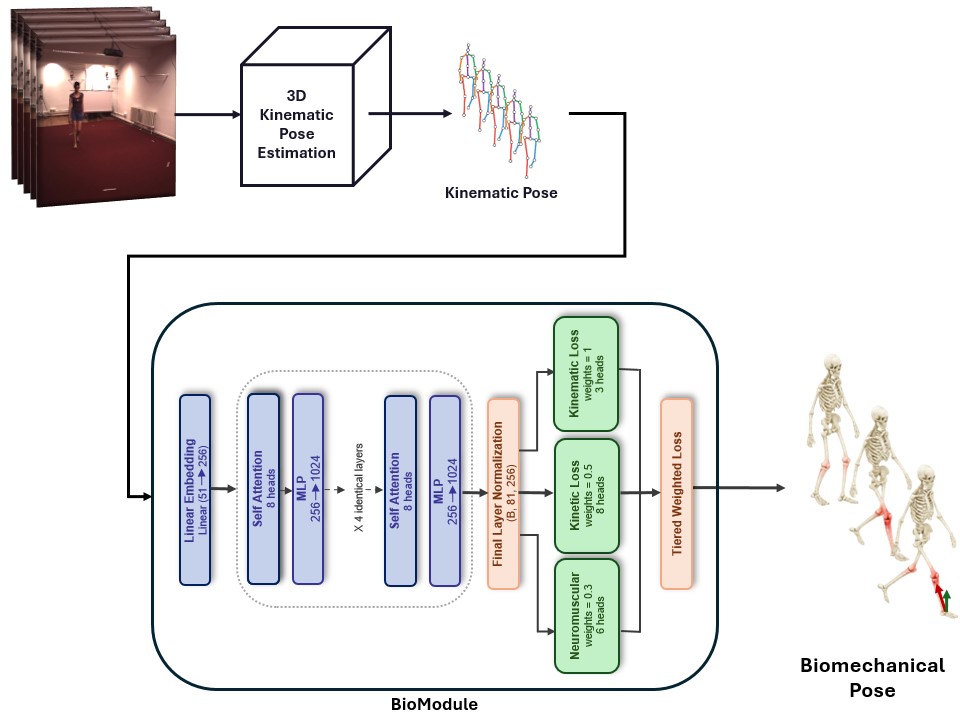}
    \vspace{-1.2em}
    \caption{End to end vision to biomechanics pipeline and BioModule architecture. A monocular video sequence is first converted into a temporally ordered 3D kinematic pose sequence by a 3D human pose estimator. BioModule then receives the root centred pose sequences, embeds the frame vectors into a latent space, and processes the temporal window with multiple layers of transformer encoders. Independent prediction heads estimate kinematic, kinetic, and neuromuscular biomechanical attributes, which are optimized using a tiered multi task loss with weights 1.0, 0.5, and 0.3, respectively.}
    \label{fig:pipeline}
    \vspace{-15mm}
\end{figure*}

We propose a pipeline that maps monocular RGB video to a biomechanical state representation of the human body through two decoupled stages, as shown in Figure~\ref{fig:pipeline}. First, a 3D human pose estimator reconstructs a temporally ordered sequence of 3D skeletal joint positions from the input video. Second, \textbf{BioModule} receives the root-centred 3D pose sequence and predicts biomechanical attributes derived from the H3.6Mplus musculoskeletal simulation labels~\cite{nasr2024using}. This separation allows BioModule to operate downstream of different 3D pose estimators without architectural changes. BioModule is trained from scratch using ground-truth 3D poses and H3.6Mplus biomechanical labels.

Let $\mathbf{P}_t \in \mathbb{R}^{J \times 3}$ denote the 3D skeleton at frame $t$, where $J{=}17$ joints follow the Human3.6M joint convention. The input to BioModule is the described sequence of Euclidean 3D joint coordinates. To remove global translation, each frame is centered at the pelvis joint $\mathbf{p}_{t,0} \in \mathbb{R}^3$:
\begin{equation}
    \bar{\mathbf{P}}_t = \mathbf{P}_t - \mathbf{1}\,\mathbf{p}_{t,0}^{\top}
    \label{eq:root_centre}
\end{equation}
The centred pose is then flattened as
$\mathbf{x}_t = \mathrm{vec}(\bar{\mathbf{P}}_t) \in \mathbb{R}^{51}$.
The resulting input window is
\begin{equation}
    \mathbf{X} = [\mathbf{x}_1,\ldots,\mathbf{x}_W]
    \in \mathbb{R}^{W \times 51}
    \label{eq:input_seq}
\end{equation}
where $W$ is the temporal receptive field. In this work, $W{=}81$, which corresponds to approximately $1.62$ seconds at 50 fps. BioModule receives raw meter scale, root centered 3D joint coordinates.

BioModule predicts $C{=}17$ biomechanical attributes
$\{\hat{\mathbf{Y}}_a\}_{a=1}^{C}$, where
$\hat{\mathbf{Y}}_a \in \mathbb{R}^{B \times W \times d_a}$ and $d_a$ denotes the output dimension of attribute $a$. These attributes are organized into three groups according to their interpretation in biomechanical musculoskeletal analysis and their relationship to vision based pose estimation, as defined in Eq.~\ref{eq:attribute_groups}.

\textbf{The kinematic attributes} describe motion geometry and its temporal derivatives. In biomechanics, kinematics refers to motion quantities without directly considering the forces that caused the motion. In human pose estimation, this group is closest to the information explicitly represented by pose sequences, because 3D joint locations encode body configuration over time. In our output set, the kinematic attributes are coordinates, speed, and acceleration. Here, there are 2 sets of coordinates: the output pose of H36M Euclidean joint positions, and the OpenSim generalized marker coordinates used in H3.6Mplus based on which the degrees of freedom for each joint are defined. Therefore, the model must learn the mapping from Human3.6M joint positions to OpenSim generalized marker coordinates as well. This mapping depends on the OpenSim skeleton, joint degrees of freedom constraints, coordinate conventions, and the anatomical model used to generate the H3.6Mplus labels.

\textbf{The kinetic attributes} describe force and load related quantities. In biomechanics, kinetics refers to the quantities associated with producing or constraining movement. In pose estimation research, these quantities are usually not predicted directly, because standard pose benchmarks evaluate geometric joint accuracy rather than physical loading. In our output set, the kinetic attributes are active torque, passive torque, ideal torque, instantaneous power, instantaneous power raw, ground reaction, seat reaction, and touch. The touch attribute is binary and represents whether or not the right and left foot have contact with floor.

\textbf{The neuromuscular attributes} describe quantities associated with excitation, activation, actuator scaling, and torque generation capacity which are derived from biomechanical simulations. The term neuromuscular in the sense is used in neuromusculoskeletal modeling, where neural excitation and activation dynamics are linked to muscle force and joint torque generation~\cite{delp2007opensim,seth2018opensim}. In our output set, the neuromuscular attributes are activation signal, excitation signal, normalized active torque, angle scaling, velocity scaling, and maximum joint torque. This grouping is used both to interpret the predicted outputs and to define the weighted multitask objective and is further used in the equations ~\ref{eq:tier_kin}, \ref{eq:tier_knt}, \ref{eq:tier_nmsc},\ and~\ref{eq:loss_total}.

\begin{equation}
\label{eq:attribute_groups}
\begin{aligned}
\mathcal{A}_{\mathrm{kin}} =
\{&
\texttt{coordinate},\ \texttt{speed},\ \texttt{acceleration}
\},\\[0.5em]
\mathcal{A}_{\mathrm{knt}} =
\{&
\texttt{active\_torque},\ \texttt{passive\_torque},\
\texttt{ideal\_torque},\ \texttt{instantaneous\_power},\\
&
\texttt{instantaneous\_power\_raw},\ \texttt{ground\_reaction},\
\texttt{seat\_reaction},\ \texttt{touch}
\},\\[0.5em]
\mathcal{A}_{\mathrm{nmsc}} =
\{&
\texttt{activation\_signal},\ \texttt{excitation\_signal},\
\texttt{normalized\_active\_torque},\\
&
\texttt{angle\_scaling},\ \texttt{velocity\_scaling},\
\texttt{maximum\_joint\_torque}
\}.
\end{aligned}
\end{equation}

\subsection{BioModule Architecture}

BioModule is a temporal transformer module that maps a sequence of root centred 3D skeletons to biomechanical attributes. The model has three main components: a framewise pose embedding, a temporal transformer encoder, and a set of independent attribute prediction heads. Figure~\ref{fig:pipeline} gives an overview of the architecture.

\subsubsection{Pose Embedding and Temporal Encoding}

Each input frame vector $\mathbf{x}_t \in \mathbb{R}^{51}$ is projected into a hidden representation using a shared linear embedding:
\begin{equation}
    \mathbf{e}_t = \mathbf{W}_e\,\mathbf{x}_t + \mathbf{b}_e
    \qquad \mathbf{e}_t \in \mathbb{R}^{d}
    \label{eq:embed}
\end{equation}
where $d{=}256$. Since self attention does not encode temporal order by itself, we add a fixed sinusoidal positional encoding~\cite{vaswani2017attention}:
\begin{equation}
    \mathrm{PE}(t,\,2i) =
    \sin\!\left(\frac{t}{10000^{2i/d}}\right)
    \qquad
    \mathrm{PE}(t,\,2i{+}1) =
    \cos\!\left(\frac{t}{10000^{2i/d}}\right)
    \label{eq:pe}
\end{equation}
The transformer input is then
\begin{equation}
    \mathbf{z}_t^{(0)} = \mathbf{e}_t + \mathrm{PE}(t)
    \qquad
    \mathbf{Z}^{(0)} \in \mathbb{R}^{B \times W \times d}
    \label{eq:z0}
\end{equation}
The positional encoding is fixed and introduces no additional trainable parameters.

\subsubsection{Temporal Transformer Encoder}

The temporal encoder contains $L{=}4$ transformer layers with pre layer normalization~\cite{xiong2020layer}. Each layer applies multi head self attention across the full temporal window, followed by a feed forward network. For layer $\ell$, the update is:
\begin{align}
    \tilde{\mathbf{Z}}^{(\ell)} &=
        \mathbf{Z}^{(\ell-1)} +
        \mathrm{MHA}\!\left(
        \mathrm{LN}\!\left(\mathbf{Z}^{(\ell-1)}\right)
        \right)
    \label{eq:attn} \\
    \mathbf{Z}^{(\ell)} &=
        \tilde{\mathbf{Z}}^{(\ell)} +
        \mathrm{FFN}\!\left(
        \mathrm{LN}\!\left(\tilde{\mathbf{Z}}^{(\ell)}\right)
        \right)
    \label{eq:ffn}
\end{align}
The attention block uses $h{=}8$ heads with per head dimension $d_k{=}32$. Attention is bidirectional over the full window. No causal mask is applied because BioModule predicts the biomechanical state associated with the centre frame rather than forecasting an unseen future frame. This allows the representation to use both preceding and following motion context.

The feed forward network uses a four times hidden expansion with GELU activation:
\begin{equation}
    \mathrm{FFN}(\mathbf{u}) =
        \mathbf{W}_2\,\mathrm{GELU}\!\left(
        \mathbf{W}_1\,\mathbf{u} + \mathbf{b}_1
        \right) + \mathbf{b}_2
    \label{eq:ffn_eq}
\end{equation}
where $\mathbf{W}_1 \in \mathbb{R}^{4d \times d}$ and
$\mathbf{W}_2 \in \mathbb{R}^{d \times 4d}$. Dropout with probability $p{=}0.1$ is used in the encoder. After the final transformer layer, a layer normalization operation produces the encoded sequence:
\begin{equation}
    \mathbf{H} = \mathrm{LN}\!\left(\mathbf{Z}^{(L)}\right)
    \qquad
    \mathbf{H} \in \mathbb{R}^{B \times W \times d}
    \label{eq:H}
\end{equation}

\subsubsection{Biomechanical Attribute Prediction Heads}

The encoded sequence $\mathbf{H}$ is passed to $C{=}17$ independent attribute prediction heads. Each head $g_a$ is a two layer MLP applied framewise:
\begin{equation}
    \hat{\mathbf{Y}}_a = g_a(\mathbf{H})
    \qquad
    \hat{\mathbf{Y}}_a \in \mathbb{R}^{B \times W \times d_a}
    \label{eq:heads}
\end{equation}
For each attribute, the head has the form
\[
    \mathrm{Linear}(d \to d/2)
    \to \mathrm{GELU}
    \to \mathrm{Dropout}(0.1)
    \to \mathrm{Linear}(d/2 \to d_a)
\]
The shared encoder learns a temporal representation of skeletal motion, while the separate heads allow each biomechanical attribute to learn its own mapping from that representation. Continuous outputs are trained and predicted in normalized space. At inference, they are converted back to physical units using
\begin{equation}
    \hat{y} = \tilde{y}\,\sigma_a + \mu_a
    \label{eq:denorm}
\end{equation}
where $\mu_a$ and $\sigma_a$ are the training set mean and standard deviation for the corresponding attribute dimension. The binary touch head outputs logits, and a sigmoid is applied at inference to obtain contact probabilities.

\subsection{Weighted Multi Task Loss}

BioModule is trained in a per-joint manner over 17 biomechanical attributes with different physical meanings, dimensionalities, and levels of uncertainty. A direct sum over all outputs would make the objective sensitive to the number of dimensions and noise level of each attribute. To reduce this effect, we use a tiered weighted multi task loss. Each attribute belongs to exactly one of the three biomechanical groups defined in Eq.~\ref{eq:attribute_groups}.

For each continuous attribute $a$, the loss is the mean squared error over the full output tensor:
\begin{equation}
    \mathcal{L}_a =
        \frac{1}{B\,W\,d_a}
        \sum_{b=1}^{B}
        \sum_{t=1}^{W}
        \sum_{j=1}^{d_a}
        \left(
        \hat{y}_{b,t,j}^{a} - y_{b,t,j}^{a}
        \right)^2
    \label{eq:mse}
\end{equation}
For the binary foot contact attribute, we use binary cross entropy with logits:
\begin{equation}
    \mathcal{L}_{\texttt{touch}} =
        \mathrm{BCEWithLogits}\!\left(
        \hat{\mathbf{Y}}_{\texttt{touch}}
        \mathbf{Y}_{\texttt{touch}}
        \right)
    \label{eq:bce}
\end{equation}

The three group losses are computed by averaging the individual attribute losses within each group:
\begin{align}
    \bar{\mathcal{L}}_{\mathrm{kin}} &=
        \frac{1}{|\mathcal{A}_{\mathrm{kin}}|}
        \sum_{a\in\mathcal{A}_{\mathrm{kin}}}
        \mathcal{L}_a
    \label{eq:tier_kin} \\
    \bar{\mathcal{L}}_{\mathrm{knt}} &=
        \frac{1}{|\mathcal{A}_{\mathrm{knt}}|}
        \sum_{a\in\mathcal{A}_{\mathrm{knt}}}
        \mathcal{L}_a
    \label{eq:tier_knt} \\
    \bar{\mathcal{L}}_{\mathrm{nmsc}} &=
        \frac{1}{|\mathcal{A}_{\mathrm{nmsc}}|}
        \sum_{a\in\mathcal{A}_{\mathrm{nmsc}}}
        \mathcal{L}_a
    \label{eq:tier_nmsc}
\end{align}

The total training objective is:
\begin{equation}
    \mathcal{L}_{\mathrm{total}} =
        1.0\,\bar{\mathcal{L}}_{\mathrm{kin}}
        + 0.5\,\bar{\mathcal{L}}_{\mathrm{knt}}
        + 0.3\,\bar{\mathcal{L}}_{\mathrm{nmsc}}
    \label{eq:loss_total}
\end{equation}

The weights reflect the relevance of each group to the observed pose sequence. Kinematic attributes receive the highest weight because they are most directly constrained by skeletal motion. Kinetic attributes receive an intermediate weight because torques, powers, and reaction forces depend on inverse dynamics, contact assumptions, and pose quality. Neuromuscular attributes receive the lowest weight because excitation, activation, and actuator level quantities are the most indirect and model dependent. Averaging within each group before applying the group weight prevents high dimensional outputs, such as the 68 dimensional neuromuscular attributes, from dominating the objective simply because they contain more output dimensions.

The loss is computed over all $W$ frames in the input window during training. During evaluation, metrics are computed only at the center frame
$t^{\star} = \lfloor W/2 \rfloor$. This avoids boundary effects and gives each reported prediction symmetric temporal context.


\section{Experimental Setup}

\subsection{Dataset}

\noindent\textbf{Base data.}
H3.6Mplus \cite{nasr2024using} extends
Human3.6M~\cite{ionescu2014human36m} with dense, per-frame biomechanical
annotations derived from subject-specific OpenSim musculoskeletal simulations.
The underlying motion capture corpus provides multi-camera video, 2D keypoint
annotations, and 3D pose data for 7~subjects (S1, S5, S6, S7, S8, S9, S11)
performing 30~standardized activities in a controlled laboratory environment,
recorded at 50\,fps across 4~synchronized camera views; the 3D joint set spans
32~body landmarks in the H36M convention.  Layered on top of this skeleton data,
H36Mplus pairs every frame with the outputs of subject-specific
inverse kinematics and inverse dynamics pipelines run in OpenSim using a
sex-matched musculoskeletal model (female: S1, S5, S7; male: S6, S8, S9, S11).
These simulations provide per-frame estimates of generalized joint coordinates,
velocities, and accelerations, active and passive joint torques, ideal torques, instantaneous mechanical power both filtered and raw,
bilateral ground reaction forces and torques, vertical seat
reaction force, binary foot contact labels, and neuromuscular signals comprising
muscle activation, neural excitation, force-length scaling, force-velocity
scaling, and maximum isometric torque, all for 68~actuators spanning 34~actuated
degrees of freedom (J7-J40). The degrees of freedom represent anatomical movements of their respective joint such as abduction, adduction, flexion, extension, etc.

\noindent\textbf{Cross-modal alignment.}
H36M world coordinates and OpenSim body-frame coordinates occupy entirely
different reference systems: H36M joints live in a camera-rig world frame, while OpenSim generalized coordinates use a musculoskeletal
body frame, making direct algebraic alignment
impossible.  We resolve this by using the H36M camera calibration matrices namely intrinsics, rotation, translation as a shared projection target,
independently projecting both H36M 3D joints and OpenSim K-markers into the same
four camera image planes.  Co-registration is verified geometrically: the
17-joint H36M subset reproduced from projected 3D coordinates matches the native
2D keypoint annotations to sub-pixel accuracy ($<0.28$\,px), and the OpenSim
pelvis marker (K1) coincides with H36M joint~0 to machine precision across all
frames and cameras.  This shared pelvis anchor ties the two coordinate systems
together frame by frame, enabling reliable cross-modal supervision.  This alignment produces frame-level paired pose and biomechanical labels for each subject-activity sequence.

\noindent\textbf{Data split.}
The dataset comprises 520{,}509 frames across 210~subject--activity clips.
Subjects S1, S5, S6, S7, and S8 (157~clips, ${\approx}108$\,min) form the
training split; subjects S9 and S11 (53~clips, ${\approx}44$\,min) are held out
as the test split and are never used during any training phase.
Per-dimension Z-score statistics ($\mu$, $\sigma$) are computed exclusively from
training frames and applied identically across all experimental phases.

\subsection{Implementation Details}

\noindent\textbf{Hardware and framework.}
All experiments are implemented in PyTorch and trained on a single NVIDIA GeForce RTX~4080 GPU. The DataLoader uses 4~worker processes with pinned memory for accelerated host-to-device transfer.

\noindent\textbf{Input preprocessing.}
Given a clip of raw 3D joint positions, we extract the 17-joint H36M subset, apply root-centring (Eq.~\ref{eq:root_centre}), flatten each frame to $\mathbf{x}_t \in \mathbb{R}^{51}$, and tile into receptive-field sequences of $W$ frames. During training, sequences are sampled with stride~1, yielding dense overlapping samples; during validation and testing, non-overlapping stride-$W$ sampling is used. Edge frames at clip boundaries are handled by edge-replication padding. Continuous biomechanical attributes are Z-score normalized per output dimension using statistics computed only from the training subjects, and the same statistics are reused during frozen evaluation and pose-estimator adaptation. Input pose coordinates are not normalized, so BioModule learns a mapping from raw, root-centered 3D skeleton geometry to normalized OpenSim-derived biomechanical attributes.

\noindent\textbf{Training setup.}
The base model is trained from random initialization for 50~epochs using AdamW~\cite{loshchilov2019decoupled} ($\beta_1{=}0.9$, $\beta_2{=}0.999$, weight decay $10^{-4}$) with an initial learning rate of $10^{-4}$ annealed via cosine scheduling~\cite{loshchilov2017sgdr} ($T_{\max}{=}50$, $\eta_{\min}{=}10^{-6}$), batch size~64, dense stride-1 sampling, and global gradient norm clipping at $1.0$. Fine-tuning from the base checkpoint runs for 10~epochs at a fixed learning rate of $10^{-5}$ with all other hyperparameters unchanged and no validation set; the best checkpoint per estimator is selected by minimum training loss.

\subsection{Evaluation Metrics}
Predictions are extracted at the center frame
$t^{\star} = \lfloor W/2 \rfloor$ of each sequence.
Continuous outputs are de-normalized via $\hat{y} = \tilde{y}\cdot\sigma + \mu$;
\texttt{touch} logits are converted to contact probabilities via sigmoid.
We report mean absolute error (MAE), root mean squared error (RMSE), and
normalized MAE (nMAE) as a percentage of each criterion's ground-truth range
over the test set for continuous attributes s, and per-foot classification accuracy
for touch. Tier-level MAE is the unweighted mean of per-attribute MAEs
within each tier.

\subsection{Benchmarked Pose Estimators}
Seven contemporary 3D kinematic pose estimators spanning dilated convolution, transformer,
diffusion, state-space, and graph architectures serve as upstream suppliers:
VideoPose3D~\cite{pavllo20193d} (2019), MHFormer~\cite{li2022mhformer} (2022),
D3DP~\cite{shan2023d3dp} (2023), PoseMamba~\cite{huang2025posemamba} (2024),
MotionAGFormer~\cite{mehraban2024motionagformer} (2024), KTPFormer~\cite{peng2024ktpformer} (2024),
and TCPFormer~\cite{liu2025tcpformer} (2025).  For each estimator, pre-extracted 3D
pose predictions are used as drop-in replacements for ground-truth poses without
post-processing, isolating BioModule's contribution from estimator-specific
design choices.

\subsection{Evaluation Protocol}
After training on ground-truth 3D poses, the learned BioModule weights are evaluated in a frozen setting by replacing the ground-truth pose input with the 3D output of each pose estimator. In this setting, no weight updates are performed. This measures how biomechanical prediction quality changes when the input skeletons come from estimated 3D poses rather than ground-truth poses. In the adaptation setting, a separate BioModule checkpoint is fine-tuned for each pose estimator using that estimator's predicted poses on the training subjects. Fine-tuning starts from the ground-truth-trained BioModule weights and runs for 10 epochs at a learning rate of $10^{-5}$. Continuous metrics are computed after de-normalization, so errors are expressed in the original physical units of each attribute. For touch, logits are converted to probabilities using a sigmoid and evaluated as binary contact predictions. Although BioModule predicts outputs for all frames in a window, evaluation uses only the center-frame prediction.
The implementation of the whole pipeline can be found at: \url{https://github.com/UTSA-VIRLab/BioModule}

\section{Results}
\subsection{Quantitative Results} 

\begin{table*}[t]
\centering
\caption{%
  Biomechanical attributes' MAE with the frozen weights protocol (RF\,=\,81),
  evaluated on subjects S9 and S11 from Human3.6M.
  The testing on the ground-truth 3D poses obviously plays as the upper bound to those of 3D poses elicited from pose estimation models.}
\label{tab:rf81_frozen}
\footnotesize
\setlength{\tabcolsep}{2.5pt}

\begin{tabular}{l ccc ccccccc}
\toprule
& \multicolumn{3}{c}{\textsc{Kinematic}}
& \multicolumn{7}{c}{\textsc{Kinetic}} \\
\cmidrule(lr){2-4} \cmidrule(lr){5-11}
\textbf{3D Pose / Model}
& Coord.  & Speed   & Accel.
& Act.T.  & Pass.T. & Ideal T. & Inst.P. & Inst.P.$_r$ & GRF   & Seat R. \\
\midrule
H36M GT
  & 0.228 & 0.193 & 1.240
  & 9.529 & 5.452 & 4.420 & 3.516 & 7.235 & 28.900 & 21.700 \\
\midrule

MHFormer
  & 0.658 & 0.403 & 2.377
  & \textbf{15.400} & 6.927 & \textbf{10.600} & 4.502 & \underline{7.524} & 39.000 & 68.400 \\

TCPFormer
  & \underline{0.534} & 0.587 & 4.290
  & 16.900 & \underline{6.536} & 12.300 & 7.002 & 8.080 & \textbf{37.500} & \textbf{42.900} \\

PoseMamba
  & 0.655 & \textbf{0.328} & \textbf{1.938}
  & \underline{16.000} & 7.008 & \underline{10.800} & \textbf{4.120} & 8.047 & \underline{38.600} & 105.400 \\

VideoPose3D
  & 0.661 & \underline{0.397} & 2.344
  & \textbf{15.400} & 6.953 & \textbf{10.600} & \underline{4.476} & 7.704 & 38.900 & 68.000 \\

MotionAGFormer
  & \textbf{0.530} & 0.594 & 4.318
  & 16.900 & \textbf{6.515} & 12.300 & 7.037 & 8.062 & \textbf{37.500} & \underline{43.000} \\

KTPFormer
  & 0.658 & 0.399 & \underline{2.343}
  & \textbf{15.400} & 6.939 & \textbf{10.600} & 4.484 & 7.537 & 39.100 & 67.100 \\

D3DP
  & 0.656 & 0.400 & 2.346
  & \textbf{15.400} & 6.937 & \textbf{10.600} & 4.500 & \textbf{7.512} & 39.100 & 66.600 \\
\bottomrule
\end{tabular}

\vspace{12pt}

\begin{tabular}{l cccccc c}
\toprule
& \multicolumn{6}{c}{\textsc{Neuromuscular}}
& \textsc{Binary} \\
\cmidrule(lr){2-7} \cmidrule(lr){8-8}
\textbf{3D Pose / Model}
& Act.Sig.& Exc.Sig.& N.Act.T. & Ang.Sc. & Vel.Sc.     & MaxJT.
& Touch \\
\midrule
H36M GT
  & 0.058 & 0.056 & 0.057 & 0.054 & 0.117 & 2.049
  & 67.500\% \\
\midrule

MHFormer
  & 0.189 & 0.124 & 0.101 & 0.217 & 0.187 & \underline{16.100}
  & \textbf{43.700\%} \\

TCPFormer
  & 0.174 & 0.168 & \underline{0.090} & 0.150 & 0.225 & \textbf{14.300}
  & 41.800\% \\

PoseMamba
  & \textbf{0.159} & \textbf{0.089} & 0.103 & 0.212 & \textbf{0.175} & 16.200
  & 42.400\% \\

VideoPose3D
  & 0.190 & 0.124 & 0.101 & 0.218 & \underline{0.186} & 16.200
  & \underline{43.600\%} \\

MotionAGFormer
  & \underline{0.174} & 0.167 & \textbf{0.090} & \textbf{0.150} & 0.226 & \textbf{14.300}
  & 42.100\% \\

KTPFormer
  & 0.190 & 0.124 & 0.101 & 0.218 & 0.186 & \underline{16.100}
  & \textbf{43.700\%} \\

D3DP
  & 0.190 & \underline{0.124} & 0.100 & 0.218 & 0.186 & \underline{16.100}
  & \underline{43.600\%} \\
\bottomrule
\end{tabular}
\end{table*}

\begin{table*}[!ht]
\centering
\caption{%
  Biomechanical attributes' MAE with the fine-tuned protocol (RF\,=\,81),
  evaluated on S9 and S11 from Human3.6M.
  Fine-tuning done through 10 epochs on the estimator's training subject poses (S1--S8)
  at $\text{lr}=10^{-5}$.
}
\label{tab:rf81_finetuned}
\footnotesize
\setlength{\tabcolsep}{2.5pt}

\begin{tabular}{l ccc ccccccc}
\toprule
& \multicolumn{3}{c}{\textsc{Kinematic}}
& \multicolumn{7}{c}{\textsc{Kinetic}} \\
\cmidrule(lr){2-4} \cmidrule(lr){5-11}
\textbf{Model}
& Coord.  & Speed   & Accel.
& Act.T.  & Pass.T. & Ideal T. & Inst.P. & Inst.P.$_r$ & GRF   & Seat R. \\
\midrule
MHFormer
  & 0.274 & 0.282 & 1.724
  & \underline{10.600} & 5.556 & 5.523 & 3.892 & 6.872 & 30.800 & 27.000 \\

TCPFormer
  & \textbf{0.260} & 0.282 & 1.689
  & \textbf{10.500} & \underline{5.532} & 5.498 & \underline{3.825} & 9.516 & \textbf{29.700} & 25.300 \\

PoseMamba
  & 0.269 & 0.276 & 1.703
  & 10.700 & 5.545 & 5.560 & 3.842 & \textbf{6.779} & 31.000 & 27.300 \\

VideoPose3D
  & 0.278 & 0.283 & 1.730
  & \underline{10.600} & 5.546 & 5.538 & 3.903 & 6.999 & 30.900 & 25.900 \\

MotionAGFormer
  & \underline{0.261} & 0.282 & \underline{1.688}
  & \textbf{10.500} & \textbf{5.524} & 5.475 & \textbf{3.816} & 9.590 & \underline{29.900} & \underline{24.800} \\

KTPFormer
  & 0.266 & \underline{0.271} & 1.690
  & \textbf{10.500} & 5.537 & \underline{5.405} & 3.857 & 6.921 & 31.100 & 25.900 \\

D3DP
  & 0.266 & \textbf{0.270} & \textbf{1.679}
  & \textbf{10.500} & 5.540 & \textbf{5.355} & 3.840 & \underline{6.862} & 30.600 & \textbf{24.200} \\
\bottomrule
\end{tabular}

\vspace{12pt}

\begin{tabular}{l cccccc c}
\toprule
& \multicolumn{6}{c}{\textsc{Neuromuscular}}
& \textsc{Binary} \\
\cmidrule(lr){2-7} \cmidrule(lr){8-8}
\textbf{Model}
& Act.Sig.& Exc.Sig.& N.Act.T. & Ang.Sc. & Vel.Sc.     & MaxJT.
& Touch \\
\midrule
MHFormer
  & 0.069 & 0.062 & \underline{0.064} & 0.070 & 0.159 & 4.635
  & 67.600\% \\

TCPFormer
  & 0.071 & 0.067 & 0.065 & 0.070 & 0.167 & 5.809
  & \textbf{68.600\%} \\

PoseMamba
  & 0.069 & 0.062 & 0.064 & 0.070 & 0.158 & 4.791
  & 67.600\% \\

VideoPose3D
  & 0.069 & 0.063 & \underline{0.064} & 0.070 & 0.160 & 6.136
  & \underline{68.300\%} \\

MotionAGFormer
  & 0.071 & 0.067 & 0.065 & 0.070 & 0.167 & 5.026
  & 68.200\% \\

KTPFormer
  & \underline{0.068} & \underline{0.061} & \textbf{0.063} & \textbf{0.067} & \underline{0.156} & \textbf{3.874}
  & 67.500\% \\

D3DP
  & \textbf{0.067} & \textbf{0.061} & \textbf{0.063} & \underline{0.067} & \textbf{0.155} & \underline{4.278}
  & \textbf{68.600\%} \\
\bottomrule
\end{tabular}

\end{table*}

The quantitative results indicates the influence of the quality and temporal consistency of the upstream 3D pose sequence on biomechanical prediction accuracy. Across the evaluated models, BioModule generally produces more reliable estimates when the input skeletons preserve anatomically plausible joint relationships and stable temporal motion patterns. This trend is expected because the predicted biomechanical variables are not independent frame-level labels. They are consequences of coordinated motion over time. Therefore, even when two pose estimators have similar average joint-position errors, their downstream biomechanical predictions may differ if one estimator produces smoother trajectories, more consistent limb orientations, or fewer local joint distortions.

The results also indicate that errors do not propagate uniformly across all biomechanical targets. Kinematic-related outputs are generally more directly tied to the observed skeletal geometry and therefore tend to be more stable across pose-estimator inputs. In contrast, kinetic and neuromuscular quantities are more sensitive to subtle changes in joint angle, velocity, and temporal coordination. Thus, the performance gap among upstream pose estimators becomes more visible as the target variable moves from geometric motion description toward physically and physiologically interpretable quantities.

The comparison across the seven pose estimators further demonstrates that pose-estimation accuracy alone is not sufficient to fully explain biomechanical prediction quality. A pose estimator that performs well in terms of joint localization may still introduce errors that are biomechanically meaningful, such as inconsistent knee flexion, unstable hip orientation, or unnatural ankle positioning during walking. These errors may have limited impact on conventional pose metrics but can strongly affect torque and muscle-related predictions. This finding supports the central motivation of BioModule: biomechanical evaluation requires attention not only to where the joints are located, but also to whether the estimated motion preserves physically meaningful relationships among body segments.

Another important observation is that BioModule remains functional across all evaluated upstream models, which supports its estimator-agnostic design. Since BioModule operates on standard 17-joint 3D skeletons, it does not require retraining or redesigning the original pose-estimation models. This makes the framework practical for comparing different pose estimators. At the same time, the variation in performance across models shows that modularity does not eliminate the influence of upstream error. Instead, the downstream structure of this research makes that influence measurable. The results therefore provide not only a benchmark of BioModule performance, but also an analysis of how pose-estimation quality propagates into biomechanical inference.

\graphicspath{{figures/qualitative/}}

\newcommand{\biocellwidth}{0.108\textwidth}

\newcommand{\biocoltitle}[1]{%
\parbox[c]{\biocellwidth}{%
\centering
\fontsize{5}{5.5}\selectfont\textbf{#1}%
}%
}

\newcommand{\biorowtitle}[1]{%
\rotatebox[origin=c]{90}{%
\parbox[c]{2.6cm}{%
\centering
\footnotesize\textbf{#1}%
}%
}%
}

\newcommand{\biocell}[2]{%
\begingroup
\setlength{\fboxsep}{1.5pt}%
\setlength{\fboxrule}{0.3pt}%
\fcolorbox{gray!35}{gray!08}{%
\parbox[c]{\dimexpr\biocellwidth-2\fboxsep-2\fboxrule\relax}{%
\centering
\vspace{1pt}
\fontsize{5}{5.5}\selectfont\textbf{#1}\\[2pt]
\includegraphics[width=\linewidth]{#2}
\vspace{1pt}
}%
}%
\endgroup
}

\begin{figure*}[htbp]
\centering
\setlength{\tabcolsep}{1.5pt}
\renewcommand{\arraystretch}{1.0}

\begin{tabular}{
>{\centering\arraybackslash}m{0.70cm}
*{8}{>{\centering\arraybackslash}m{\biocellwidth}}}

&
\biocoltitle{GT} &
\biocoltitle{D3DP} &
\biocoltitle{KTPFormer} &
\biocoltitle{MHFormer} &
\biocoltitle{MotionAG\\Former} &
\biocoltitle{PoseMamba} &
\biocoltitle{TCPFormer} &
\biocoltitle{VideoPose3D}
\\[6pt]

\biorowtitle{Active Torque} &
\biocell{7.21}{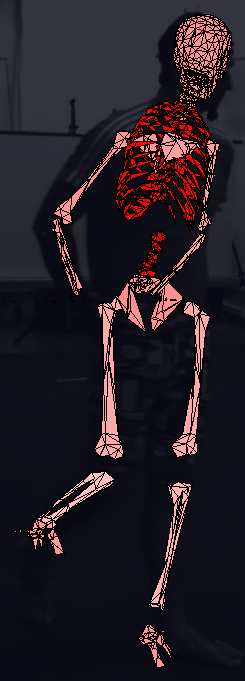} &
\biocell{10.24}{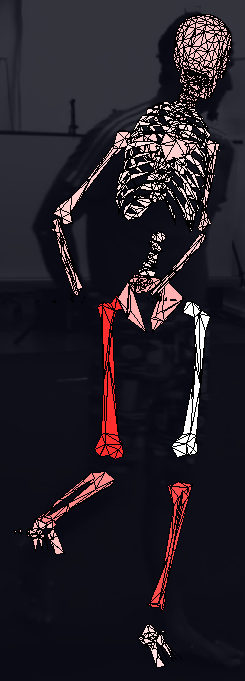} &
\biocell{9.23}{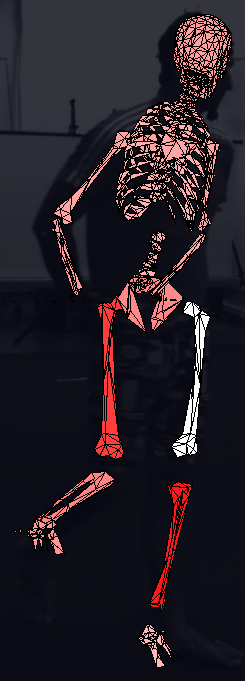} &
\biocell{9.18}{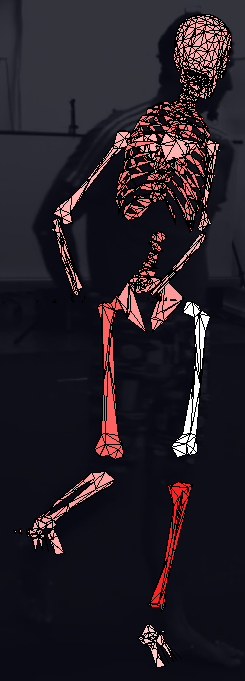} &
\biocell{5.20}{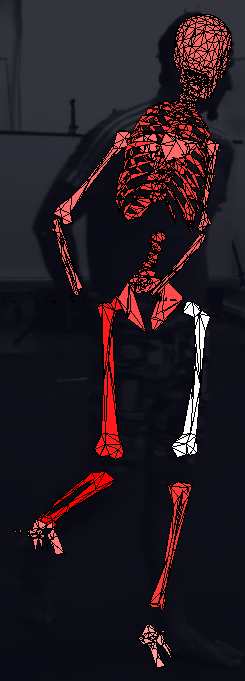} &
\biocell{4.97}{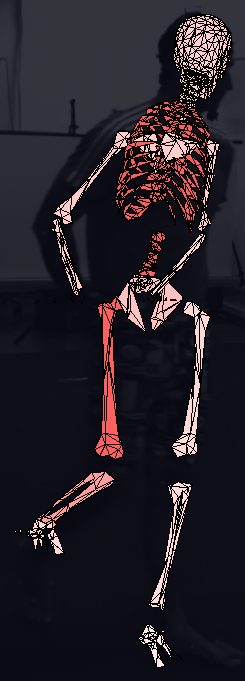} &
\biocell{1.60}{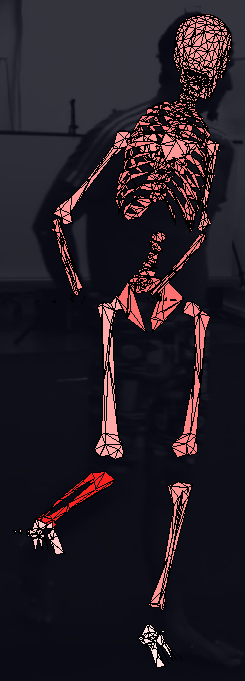} &
\biocell{9.91}{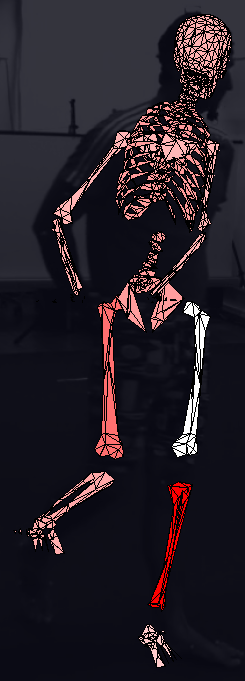}
\\[6pt]

\biorowtitle{Passive Torque} &
\biocell{0.65}{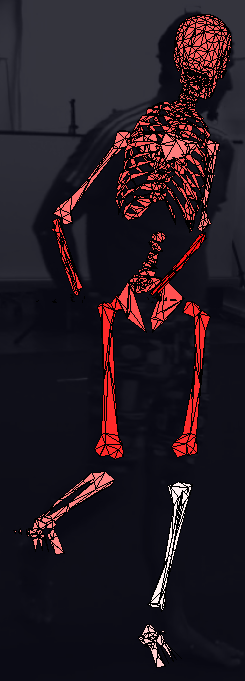} &
\biocell{1.00}{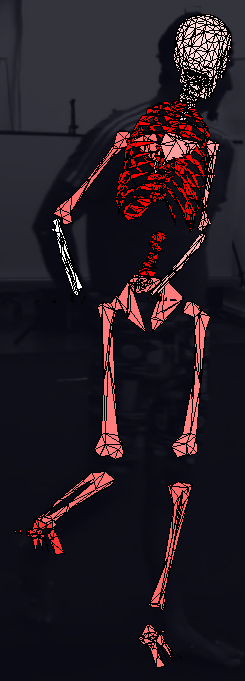} &
\biocell{0.96}{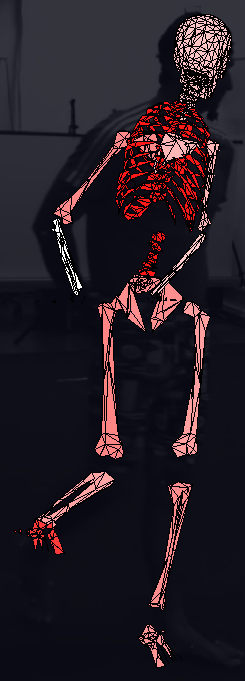} &
\biocell{1.24}{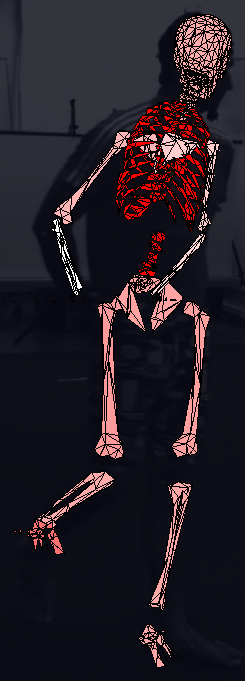} &
\biocell{0.94}{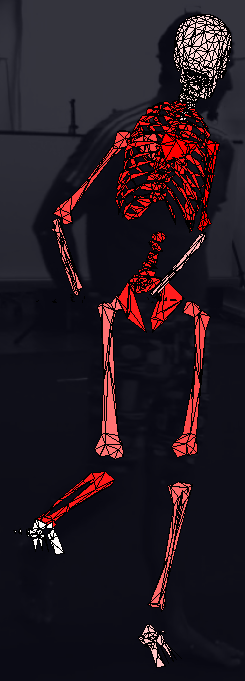} &
\biocell{1.44}{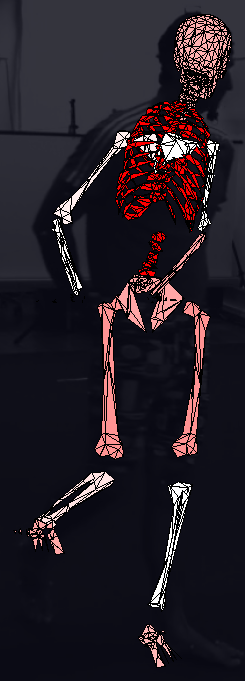} &
\biocell{0.92}{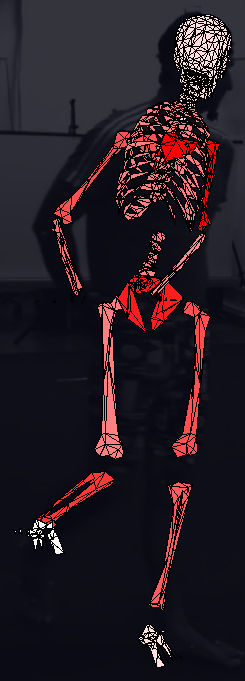} &
\biocell{1.07}{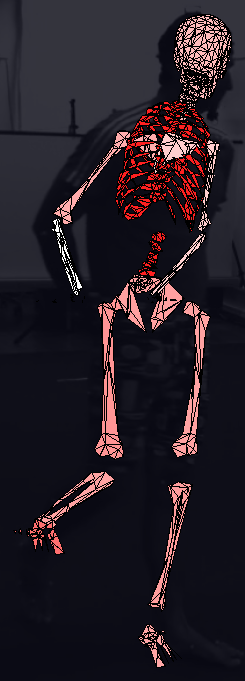}
\\[3pt]

\biorowtitle{Muscle Activation} &
\biocell{0.06}{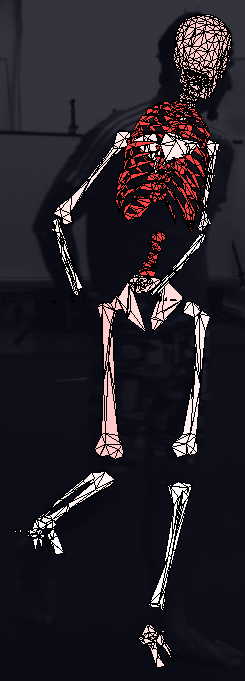} &
\biocell{0.14}{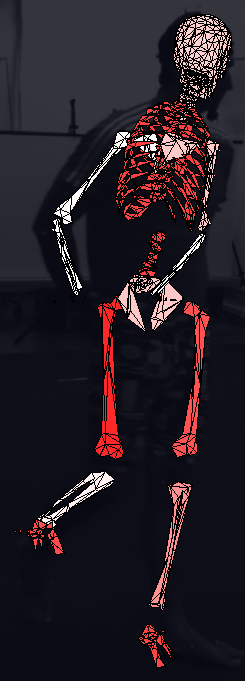} &
\biocell{0.14}{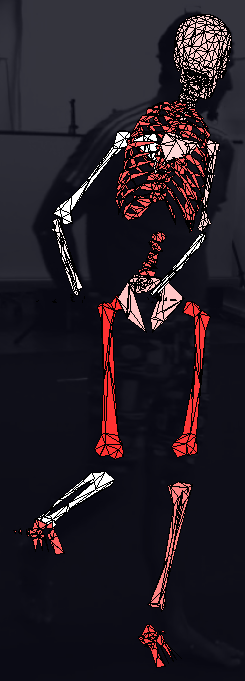} &
\biocell{0.13}{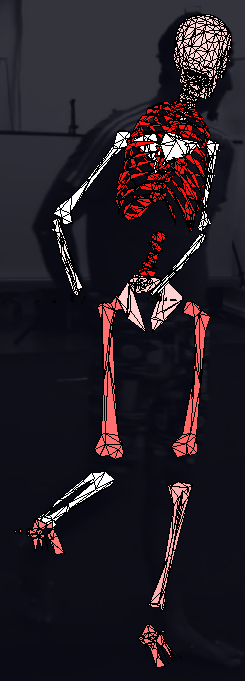} &
\biocell{0.17}{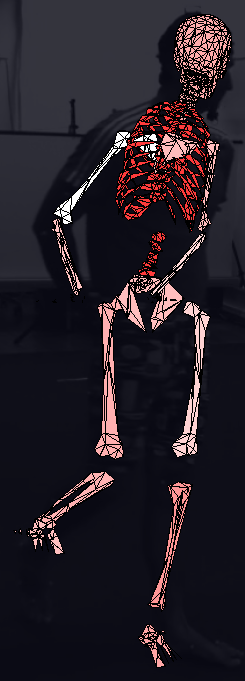} &
\biocell{0.09}{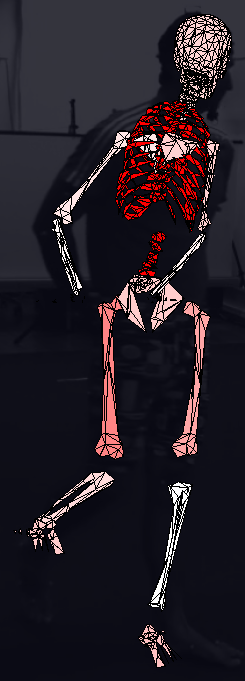} &
\biocell{0.15}{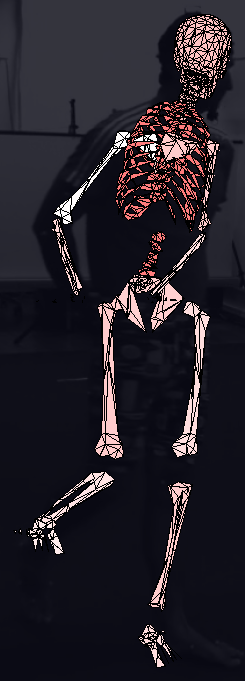} &
\biocell{0.13}{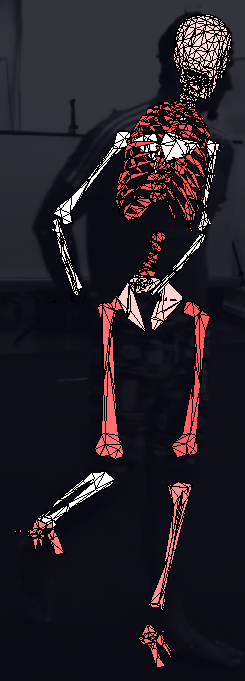}
\\[3pt]

\biorowtitle{Neural Excitation} &
\biocell{0.09}{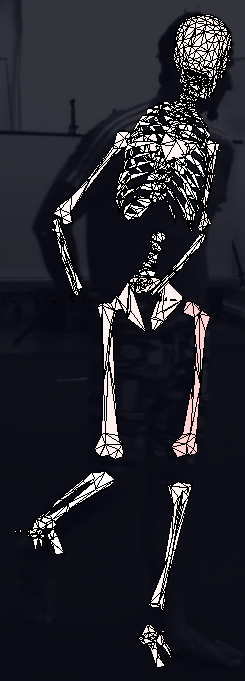} &
\biocell{0.03}{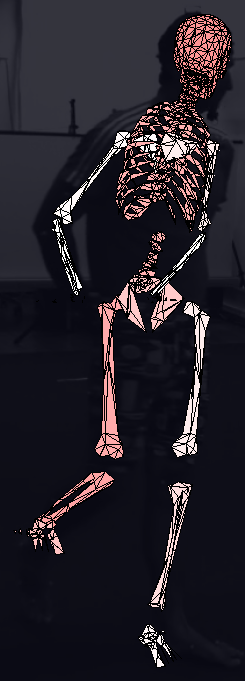} &
\biocell{0.03}{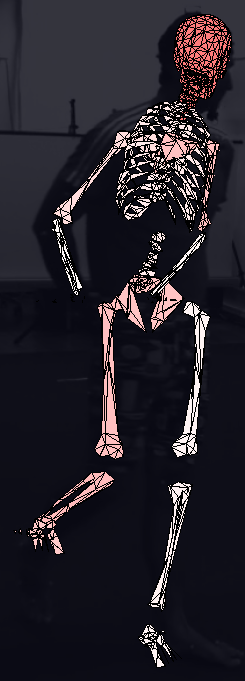} &
\biocell{0.04}{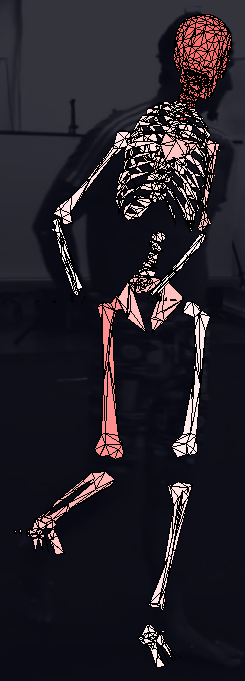} &
\biocell{0.11}{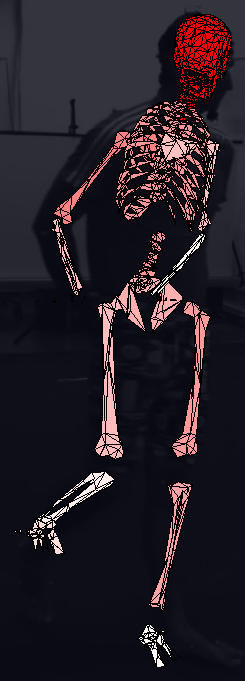} &
\biocell{0.05}{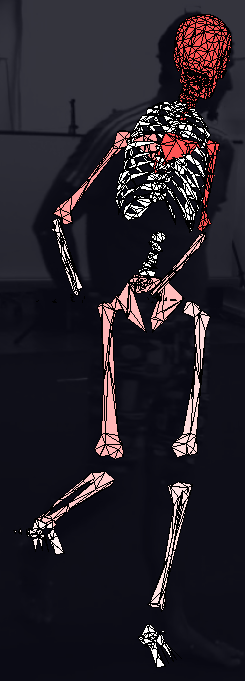} &
\biocell{0.11}{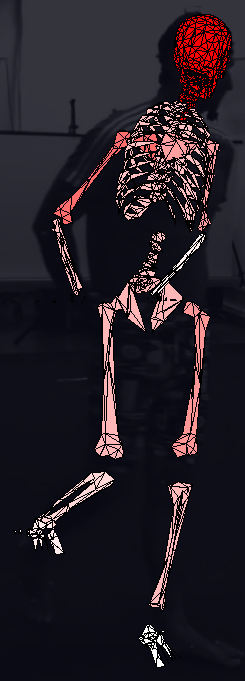} &
\biocell{0.03}{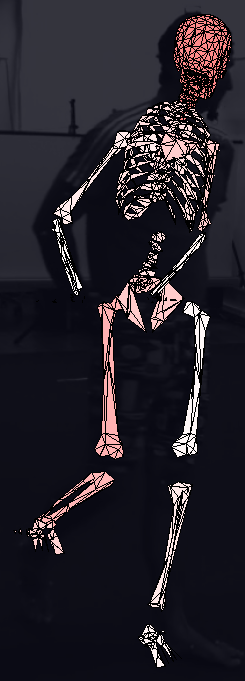}
\\

\end{tabular}

\caption{Qualitative comparison of BioModule predictions for a sample walking frame from test subject S9. Rows show four biomechanical attributes. Columns compare predictions obtained from ground-truth 3D poses and from seven upstream 3D pose estimators. Bone coloring encodes the predicted attribute magnitude and the number above each panel reports the mean predicted value across all body segments.}
\label{fig:bio_grid}
\end{figure*}

\subsection{Qualitative Results}
Skeletal visualizations of four biomechanical attributes from test subject S9 during a walking sequence are presented in Figure~\ref{fig:bio_grid}. The figure compares the BioModule outputs obtained from ground-truth 3D poses and from seven upstream 3D pose estimators for the same representative frame. The red spectrum indicates the value magnitude of the corresponding attribute across body segments.

The walking frame provides an informative case because gait involves coordinated loading across the hip, knee, and ankle, with different muscles and torques contributing at different phases of the movement.In the ground-truth visualization, higher biomechanical responses are concentrated around the lower-limb joints, particularly the hip, knee, and ankle, reflecting their dominant role in generating and controlling walking motion. This spatial distribution serves as a reference for assessing whether the predicted biomechanical patterns remain anatomically plausible. 

For active torque, the qualitative results are expected to reveal how well each pose-estimator input preserves the joint-level loading pattern of the walking frame. Active torque reflects the net muscular effort required to generate or control motion. Models that produce less accurate pose input with local distortions around the lower limbs may shift the predicted torque magnitude or produce unnatural concentration at the wrong joint.

Unlike active torque, passive torque is strongly related to joint configuration and soft-tissue resistance and hence more sensitive to joint-angle errors. If an upstream pose estimator produces excessive or insufficient knee flexion in the walking frame, the predicted passive torque may become exaggerated or suppressed relative to the ground truth. This makes passive torque a useful qualitative indicator of whether the estimated skeleton remains within plausible biomechanical ranges.

 Muscle activation is not directly visible from the skeleton, but it is inferred from the relationship between posture, motion, and the learned biomechanical supervision. Scattered or misplaced activation patterns may indicate that the input skeleton lacks the temporal coherence needed to support reliable muscle-level inference.

Neural excitation is expected to be among the more challenging outputs because it represents a control signal rather than a directly observable geometric quantity. Qualitative differences in neural excitation therefore provide insight into the limits of downstream inference from 3D skeletons alone. If the predicted excitation patterns remain spatially and functionally consistent with the ground truth, this supports the ability of BioModule to infer higher-level biomechanical attributes from pose sequences. If the predictions become noisy or anatomically inconsistent, this suggests that some neuromuscular quantities may require richer input representations, stronger temporal modeling, or additional physical constraints.

The qualitative results are expected to show that the best-performing upstream models do not simply produce cleaner skeletons; they preserve biomechanically meaningful structure which also depends highly on the action scenario. When the upstream pose contains local errors, the downstream biomechanical maps may amplify those errors in different manners.

The visual patterns in Figure~\ref{fig:bio_grid} support the quantitative findings. They show that the downstream formulation makes biomechanical consequences of pose-estimation error visible. Rather than treating all pose errors as equally important, the visualizations reveal which errors matter more for physical interpretation. This distinction is central to the purpose of BioModule: to connect pose-estimation outputs with biomechanical meaning.

\section{Discussion}
This study evaluates BioModule as a modular temporal transformer for predicting biomechanical attributes from 3D human pose sequences. In this work, biomechanical prediction is treated as a downstream task as a methodological design choice which allows BioModule to attach to existing 3D pose estimators without modifying their architectures. biomechanical attributes can be learned from pose sequences alone. In this sense, BioModule provides an initial bridge between the conventional 3D pose-estimation setting and biomechanical analysis of human motion. As a byproduct,this setting makes it possible to compare how different upstream pose models affect downstream biomechanical prediction.

Benchmarking of BioModule on the outputs of SOTA 3D pose estimation model sets an important milestone because most video-based human pose estimation pipelines produce sparse skeletal representations rather than full musculoskeletal states. Therefore, by learning the mapping from 3D pose sequences to simulation-derived biomechanical attributes, BioModule shows that biomechanical interpretation can be approached directly from video-compatible pose representations.

In qualitative walking visualizations, the skeletal maps of active torque, passive torque, muscle activation provide insight beyond numerical error values by showing where biomechanical demand is concentrated and whether the predicted distribution remains consistent with expected movement behavior.

BioModule provides a baseline formulation for connecting standard 3D pose estimation with biomechanical motion analysis. By predicting biomechanical attributes from pose sequences, it treats the skeleton as an intermediate representation rather than only a geometric output. This provides an initial step toward video-based biomechanical assessment without motion-capture systems, wearable sensors, force plates, or separate inverse-kinematics pipelines.

\textbf{Limitations and Future Work}: The current study is intended as a baseline rather than a complete solution for in-the-wild biomechanical analysis. BioModule relies on aligned biomechanical supervision and therefore inherits limitations from the underlying simulation-derived labels, musculoskeletal assumptions, and dataset alignment process. In addition, the use of a reduced 17-joint skeleton improves compatibility with standard 3D pose estimators but limits anatomical detail compared with full-body marker sets or subject-specific musculoskeletal models.

The evaluation is also limited to Human3.6M-based motion data, which is captured in a controlled environment and does not fully represent outdoor videos, occlusion, camera motion, clothing variation, clinical movement patterns, or complex sports activities. Therefore, the present results should be interpreted as evidence that biomechanical attributes can be learned from pose sequences under controlled conditions, and more data including pose and the corresponding biomechanical attributes is needed to bring the field to the point where validation for unconstrained real-world deployment is feasible.

Future work should extend this framework to richer skeletal representations, subject-specific biomechanical modeling, uncertainty-aware prediction, and broader activity domains. Further validation is needed on real-world video, sports movements, rehabilitation tasks, and clinical populations. A longer-term direction is to combine video-based pose estimation with BioModule-like biomechanical prediction so that human motion analysis can be performed without intrusive sensors, expensive laboratory equipment, or separate inverse-kinematics pipelines.

\section{Conclusion}

This work presented BioModule, a lightweight temporal transformer for predicting biomechanical attributes from 3D human pose sequences. In this study, biomechanical prediction was formulated as a downstream modular extension of 3D pose estimation. This formulation was chosen to allow BioModule to operate on the outputs of existing pose estimators without modifying their architectures. It should therefore be understood as the design adopted in this research, not as a claim that biomechanical estimation must always be performed downstream of pose estimation.

Using aligned Human3.6M and Human3.6Mplus supervision, BioModule was trained to infer biomechanical quantities from standard 17-joint skeletons. The experiments across seven upstream pose estimators showed that biomechanical prediction is strongly affected by the anatomical plausibility and temporal consistency of the input pose sequence. The results also showed that different biomechanical outputs have different sensitivity to upstream error. Kinematic quantities are more directly related to skeletal geometry, while kinetic and neuromuscular quantities, including active torque, passive torque, muscle activation, and neural excitation, are more sensitive to subtle pose and motion artifacts.

The qualitative walking visualizations further demonstrated that biomechanical evaluation provides insight beyond conventional pose-estimation metrics. Errors that may appear small geometrically can become important when they occur near mechanically active joints such as the knee, hip, or ankle. By visualizing predicted biomechanical quantities across the skeleton, BioModule helps reveal whether an estimated pose sequence preserves physically meaningful movement structure.

Overall, the findings indicate that BioModule can serve as a compact bridge between computer vision-based 3D pose estimation and biomechanical motion analysis. Its estimator-agnostic design makes it useful for extending existing pose-estimation pipelines, while its output space provides a more functionally meaningful way to evaluate human motion. Future work will explore richer skeletal representations, uncertainty-aware biomechanical prediction, stronger physical constraints, and broader validation across activities, subjects, and real-world movement conditions.

\section{Acknowledgments}
This material is partially based upon work supported by the National Science Foundation under Grant No. 2153249. Any opinions, findings, and conclusions or recommendations expressed in this material are those of the author(s) and do not necessarily reflect the views of the National Science Foundation.

{
\bibliographystyle{sn-mathphys-num}
\bibliography{main}
}
\end{document}